\documentclass{ieeeaccess}
\usepackage{cite}
\usepackage{amsmath,amssymb,amsfonts}
\usepackage{algorithmic}
\usepackage{graphicx}
\usepackage{textcomp}
\usepackage{url}
\usepackage{booktabs}

\usepackage{bm}
\makeatletter
\AtBeginDocument{\DeclareMathVersion{bold}
\SetSymbolFont{operators}{bold}{T1}{times}{b}{n}
\SetSymbolFont{NewLetters}{bold}{T1}{times}{b}{it}
\SetMathAlphabet{\mathrm}{bold}{T1}{times}{b}{n}
\SetMathAlphabet{\mathit}{bold}{T1}{times}{b}{it}
\SetMathAlphabet{\mathbf}{bold}{T1}{times}{b}{n}
\SetMathAlphabet{\mathtt}{bold}{OT1}{pcr}{b}{n}
\SetSymbolFont{symbols}{bold}{OMS}{cmsy}{b}{n}
\renewcommand\boldmath{\@nomath\boldmath\mathversion{bold}}}
\makeatother
\renewcommand{\thetable}{\Roman{table}}

\def\BibTeX{{\rm B\kern-.05em{\sc i\kern-.025em b}\kern-.08em
    T\kern-.1667em\lower.7ex\hbox{E}\kern-.125emX}}

\begin{document}

\title{Representation Transfer of Foundation Models for Ultra-Widefield Retinal Imaging}

\author{
\uppercase{MINGYA ALEXA GONG}\authorrefmark{1},
\uppercase{DA MA}\authorrefmark{2,3},
\uppercase{LOVRE ANTONIO BUDIMIR}\authorrefmark{4},
\uppercase{IVANA MATOVINOVIC}\authorrefmark{4},
\uppercase{SVEN LONCARIC}\authorrefmark{4},
\uppercase{MYEONG JIN JU}\authorrefmark{5,6},
\uppercase{YUKUN ZHOU}\authorrefmark{1,7},
\uppercase{SIEGFRIED K. WAGNER}\authorrefmark{1,7},
\uppercase{PEARSE A. KEANE}\authorrefmark{1,7},
AND
\uppercase{MARINKO V. SARUNIC}\authorrefmark{1,8}
}

\address[1]{Institute of Ophthalmology, University College London, London, United Kingdom}

\address[2]{Wake Forest University School of Medicine, Winston-Salem, NC, USA}

\address[3]{Virginia Tech-Wake Forest University School of Biomedical Engineering and Sciences, Blacksburg, VA, USA}

\address[4]{University of Zagreb Faculty of Electrical Engineering and Computing, Zagreb, Croatia.}

\address[5]{Department of Ophthalmology and Visual Sciences, University of British Columbia, Vancouver, BC, Canada}

\address[6]{School of Biomedical Engineering, University of British Columbia, Vancouver, BC, Canada}

\address[7]{NIHR Biomedical Research Centre, Moorfields Eye Hospital NHS Foundation Trust, London, United Kingdom}

\address[8]{Department of Medical Physics and Biomedical Engineering, University College London, London, United Kingdom}

\tfootnote{This work was supported by: BRC Translational Studentship (Imaging) funded by Moorfields Biomedical Research Centre; NIHR BRC at Moorfields and UCL Institute of Ophthalmology; Moorfields Eye Charity. Dr. Ma is supported by Wake Forest Translational Eye and Vision Research Center (TrEVR) and Center for Artificial Intelligence Research (CAIR) pilot awards. Dr. Keane is supported by a UK Research {\&} Innovation Future Leaders Fellowship (MR/T019050/1), Moorfields Eye Charity with The Rubin Foundation Charitable Trust (GR001753), and an Alcon Research Institute Senior Investigator Award.}

\markboth
{Gong \headeretal: Preprint}
{Gong \headeretal: Preprint}

\corresp{Corresponding author: Mingya Alexa Gong (e-mail: alexa.gong.19@ucl.ac.uk).}

\begin{abstract}
Despite the widespread adoption of foundation models as feature extractors for medical imaging, relatively little is understood about how different pretraining strategies influence the transferability of learned representations to weakly supervised ophthalmic imaging tasks. We investigate this question in ultra-widefield (UWF) retinal imaging by evaluating foundation model representations within a patch-based multiple instance learning (MIL) framework for disease classification on UWF images. We compare Vision Transformer encoders pretrained with supervised, Masked Autoencoder (MAE), and self-distillation objectives, while keeping the downstream aggregation architecture unchanged. Within a controlled comparison of ViT-B encoders pretrained on ImageNet-1k, the choice of pretraining objective substantially influenced frozen representation transfer, with supervised and self-distillation-based models outperforming MAE. A contemporary DINOv3 model pretrained at a larger scale achieved the strongest overall performance, with a quadratic weighted kappa of 0.863 for five-class diabetic retinopathy grading, comparable with DINOv1. Attention analysis further revealed distinct patch-aggregation behaviours associated with the different pretrained representations, while partial fine-tuning substantially reduced the performance gap for MAE. These findings suggest that pretraining strategy influences both representation transferability and the subsequent aggregation of patch-level evidence within MIL, resulting in differences in downstream classification performance.
\end{abstract}


\titlepgskip=-21pt

\maketitle

\section{Introduction}
\label{sec:introduction}
\PARstart{F}{oundation} models have emerged as a powerful strategy for learning generalisable representations from large-scale imaging datasets [1]. In medical imaging, the development of accurate deep learning models is often constrained by the limited availability of labelled data, as manual annotation is costly, time-consuming, and requires domain expertise. Transfer learning has become a widely adopted approach, where pretrained foundation models can be adapted to downstream medical imaging tasks [2]. Such pretraining can provide rich representations that improve performance compared with models trained from scratch [3]. 

With the increasing availability of large collections of unlabelled data, the development of self-supervised foundation models has further accelerated, allowing models to learn meaningful representations without requiring explicit labels. Rather than optimising directly for a downstream task, self-supervised methods learn visual representations through pretext tasks such as image reconstruction, allowing the model to learn useful features that transfer to smaller downstream tasks.
Modern foundation models are trained using diverse pretraining paradigms, including supervised learning [4], reconstruction-based self-supervision [5], and self-distillation [6], [7]. Recent models such as DINOv3 [7] have further demonstrated that scaling both model capacity and pretraining data can substantially improve representation transferability across a wide range of downstream vision tasks. As a result, foundation models have become an increasingly important component of medical image analysis pipelines.

In ophthalmic imaging, several domain-specific foundation models have recently been proposed for conventional imaging modalities such as colour fundus photography (CFP) and optical coherence tomography (OCT), including RETFound [8]. However, unlike natural image analysis, ophthalmic imaging encompasses a wide range of imaging technologies and modalities, making it difficult for a single foundation model to capture the full diversity of retinal image characteristics. Consequently, researchers frequently rely on transferring representations learned from other domains, such as large-scale natural image datasets [9]. Understanding how well these representations transfer is particularly important for ultra-widefield (UWF) retinal imaging, where the large field-of-view and high image resolution present additional computational challenges. UWF Optomap imaging is an established clinical imaging technology that provides a field-of-view (FOV) of up to 200$^{\circ}$ and covers approximately 80\% of the retinal surface in a single scan [10]. Compared with conventional fundus imaging (30 - 60$^{\circ}$), this expanded field of view can offer a comprehensive assessment of the periphery, potentially allowing the discovery of new retinal biomarkers. However, UWF images typically have larger image sizes (often 4000×4000 pixels), whereas most deep learning pipelines operate on much smaller inputs (e.g. 224×224 pixels) [11]. Downsampling of UWF images may potentially discard clinically relevant information, such as microaneurysms, haemorrhages, or tumours that only occupy a few pixels. Preserving this high-resolution spatial information while maintaining computational feasibility therefore remains a key challenge for applying foundation models to UWF imaging.

Multiple instance learning (MIL) [12] provides a potential framework for addressing this challenge. MIL is a weakly supervised learning framework in which each image is treated as a \emph{bag} composed of smaller patches (\emph{instances}), allowing models to learn from local regions while preserving image-level supervision [13]. This formulation enables patch-level representations learned by foundation models to be evaluated, providing insight into how local image information is selected and aggregated to produce image-level predictions. MIL has been widely adopted in computational pathology for analysing large whole-slide images [14], [15], [16], and has recently begun to appear in ophthalmic imaging applications [17]. However, existing studies typically focus on improving aggregation mechanisms, while the role of the underlying feature extractor representation remains comparatively unexplored.

In this work, we investigate how different pretrained visual representations transfer to high-resolution UWF retinal image analysis within a controlled patch-based MIL framework. We compare generalist Vision Transformer (ViT) encoders pretrained with supervised learning, Masked Autoencoder (MAE), and self-distillation-based objectives on multiple UWF retinal classification datasets. In addition to standard classification metrics, we investigate the effect of patch-based image representation, analyse MIL attention distributions to characterise patch aggregation behaviour, and examine the impact of limited encoder adaptation through partial fine-tuning.

Our contributions are as follows:
\begin{itemize}
    \item We present a controlled evaluation of how different foundation model pretraining strategies transfer to high-resolution UWF retinal image analysis within a fixed patch-based MIL framework.
    \item We demonstrate that foundation model pretraining objectives influence downstream classification performance and attention behaviour within the MIL framework, with self-distillation-based and supervised ViTs outperforming MAE-based approaches.
    \item We provide, to our knowledge, one of the first systematic evaluations of foundation model representations within a MIL framework for UWF retinal image analysis.
\end{itemize}

\section{Related Work}
\subsection{Vision Foundation Models and Pretraining Strategies}
ViTs were first introduced by Dosovitskiy et al. [4] in 2020. By representing images as a sequence of fixed-size patches and employing self-attention mechanisms to model long-range relationships between image regions, ViTs demonstrated strong scalability and transferability across a wide range of visual recognition tasks [19].

Early ViTs were trained using supervised learning on large, labelled datasets such as ImageNet [20], achieving strong performance on downstream tasks. However, the dependence on large-scale manual annotations motivated the development of self-supervised learning approaches, capable of learning representations from much larger, unlabelled image datasets.

In addition to reducing annotation requirements, self-supervised learning enables ViTs to be pretrained on extensive datasets before being fine-tuned or deployed as frozen feature extractors for downstream applications. This often results in improved downstream performance, compared with models trained from scratch [8], [21]. Modern vision foundation models can be pretrained using a variety of self-supervised learning strategies, including generative (such as MAE) and discriminative approaches (such as self-distillation). These approaches learn representations through different supervisory signals and may consequently capture different visual characteristics. For example, MAEs [5] learn image representations through image reconstruction, whereas self-distillation methods such as DINO [6], [7] learn representations through encouraging consistency across augmented views of the same image.

\subsection{Foundation Models in Ophthalmic Imaging}
Foundation models are particularly attractive in ophthalmic imaging, where labelled datasets are often limited and expert annotation is costly. Several specialised foundation models have been developed. For example, RETFound [8] was pretrained on a large-scale CFP and OCT dataset and achieved strong transferability across multiple downstream tasks, including diabetic retinopathy (DR) classification. Subsequent studies have further explored ophthalmic foundation models through large pretraining datasets and with the incorporation of additional imaging modalities [22], [23].

However, most existing work focuses on developing specialised foundation models and improving downstream tasks, with less attention paid to understanding how different pretraining strategies and pretext tasks influence downstream performance when transferring general-purpose representations to ophthalmic imaging tasks. Understanding this relationship is particularly important when foundation models are deployed as frozen feature extractors, where downstream performance depends heavily on the transferability of the learned representations. 

\subsection{Foundation Models in Multiple Instance Learning}
MIL provides a useful framework for evaluating the transferability of foundation model representations because it relies on patch-level features extracted by a pretrained encoder prior to aggregation. MIL has been widely adopted in medical imaging as a weakly supervised framework for analysing high-resolution images, such as UWF retinal imaging, where only image-level labels are available. Recently, transformer-based architectures have been integrated into MIL frameworks as feature extractors prior to instance aggregation. As ViTs naturally operate on image patches, they are well-suited for patch-based learning paradigms [4]. For example, Yu et al. proposed ‘MIL-VT’ for DR classification, a ViT-Small model with an added ‘MIL head’ [24] that outperformed CNN models under the same setup. Similarly, Bi et al. proposed ‘MIL-ViT’ [25], a framework that integrates a ViT backbone with an attention-based MIL module to highlight local pathological regions in fundus images, demonstrating improved performance on several retinal disease classification benchmarks. More recently, Yang et al. proposed a transformer-based MIL framework for DR classification in which retinal fundus images were divided into 224×224 patches and aggregated for image-level prediction [26]. However, most existing studies evaluate a single pretrained encoder and focus primarily on downstream classification performance, providing limited insight into how different representation learning strategies affect feature aggregation within MIL.

\section{Methodology}
\subsection{Datasets}
For downstream evaluation, labelled datasets were used for disease classification experiments. Each dataset had a different scale and split availability, so the evaluation protocols were adapted accordingly. 

\subsubsection{Five-class DR (Primary analysis dataset)}
We used the publicly available MMRDR dataset, a multimodal retinal dataset comprising colour fundus photography (CFP), UWF, and optical coherence tomography (OCT) images [18]. The UWF subset uses five-class severity grading: no DR, mild, moderate, severe and proliferative DR (pDR), with labels defined by four senior ophthalmologists and a retinal specialist through a structured calibration protocol. The dataset contains 10,404 UWF images from 6,109 patients and predefined train and test sets split at the patient level, which we retained to enable standardised evaluation. We further divided the train set into training and validation sets at a 4:1 split, stratified by disease grade, for model selection and hyperparameter tuning such as early stopping. Our training set contains 2058 images for no-DR, 1142 for mild, 1298 for moderate, 815 for severe, and 934 images for pDR. We reported the performance of the trained encoder+MIL model on the unseen test set. Each experiment was repeated across three different seeds while keeping the training, validation, and test splits fixed.

\subsubsection{Non-referable/referable DR (Additional DR validation)}
We used a Deep DR Image Dataset (DeepDRiD [27]), containing 254 UWF DR images with severity grading (no/mild/moderate/severe/proliferative). Due to the severe and proliferative classes being imbalanced, we combined classes and rearranged the dataset for binary classification of non-referable (no/mild) and referable (moderate/severe/proliferative) DR, resulting in 147 non-referable images (train = 87, validation = 30, test = 30) and 107 referable images (train = 65, validation = 20, test = 22). The DeepDRiD dataset provides predefined training, validation and test splits, which were used directly in our experiments to ensure consistency with prior work. For our experiments, we used the same training and validation splits but repeated each experiment three times with different seeds to record training variance. The validation split was used for model selection and early stopping. Performance was reported on the test split.

\subsubsection{Healthy/Intraocular Tumours (Cross-disease validation)}

We used a publicly available dataset containing 2,031 UWF fundus images released by Sun et al. [28], containing five types of intraocular tumours and normal images. As the proportion of normal images to tumour images was double (healthy = 1354, with tumour = 677), we randomly sampled 600 images from each class to create a balanced subset, resulting in 600 normal and 600 intraocular tumour images. As no predefined training and test split was available, 5-fold stratified cross-validation was used to provide a lower-variance estimate than a single train/test partition. For each outer cross-validation fold, the held-out fold was used exclusively for evaluation. The remaining four folds were further divided into training (80\%) and validation (20\%) subsets, with the validation subset used for early stopping and model selection. The dataset includes multiple images from the same eye acquired during longitudinal follow-up examinations, but because patient identifiers were not available, splitting could only be performed at the image level. 

All images for the downstream tasks were resized to 1024×1024 pixels, converted to single-channel greyscale, and normalised with mean and standard deviation of 0.5 using PyTorch and torchvision transforms [29]. A common grayscale representation was used across all encoders to ensure that differences in downstream performance were not driven by differences in input channel information. This was particularly important because the domain-specific AlzEye MAE had been pretrained using grayscale UWF images (see Appendix A). For encoders expecting three-channel input, the grayscale channel was therefore replicated three times.

\subsection{Patch-based MIL Framework}
\subsubsection{Implementation}
\begin{figure*}[!t]
\centering
\includegraphics[width=\textwidth,height=0.85\textheight,keepaspectratio]{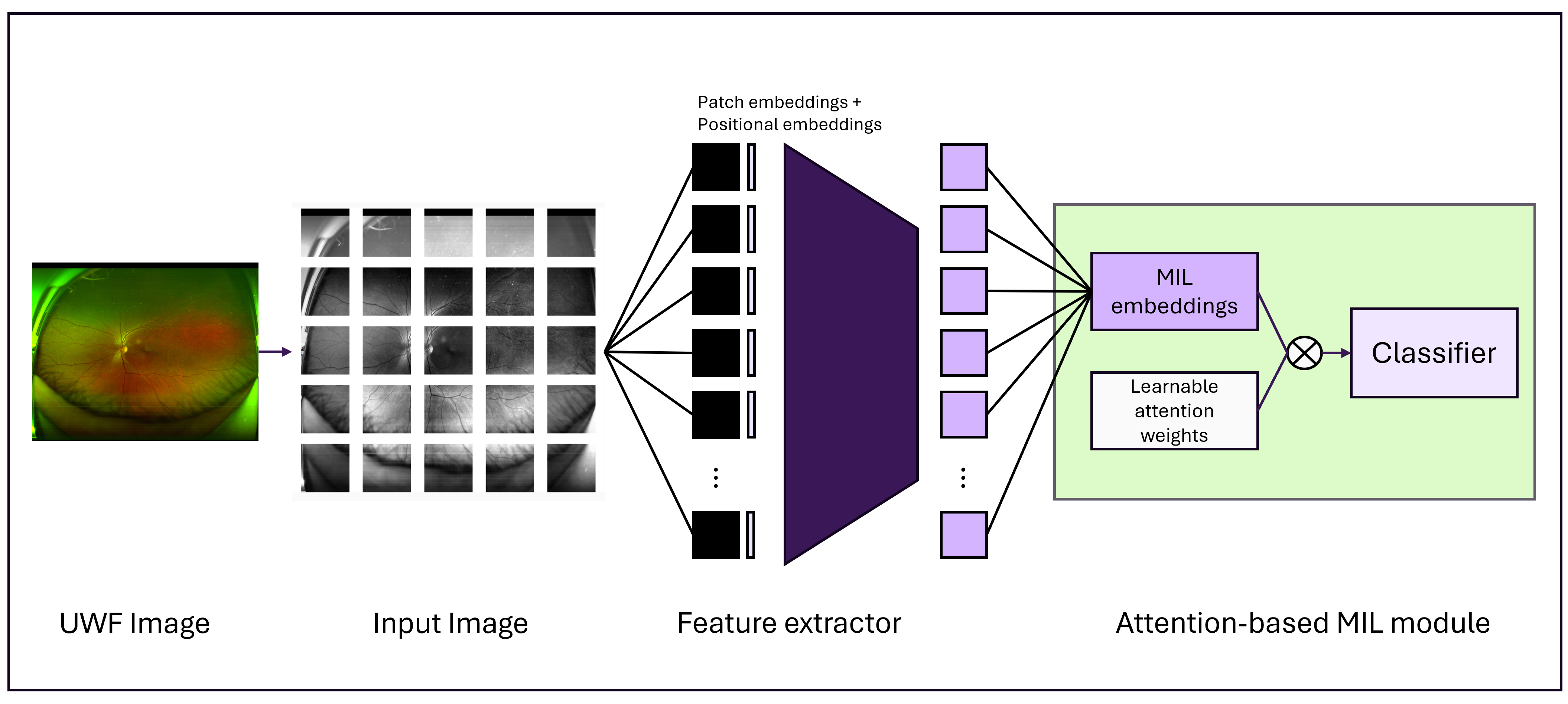}
\caption{Overview of the patch-based multiple instance learning framework used to evaluate foundation model representations. Ultra-widefield retinal images are divided into patches and processed by a feature extractor. Patch-level features are aggregated using an attention-based MIL module to produce a bag-level representation for classification.}
\label{fig1}
\end{figure*}

To provide a controlled framework for evaluating foundation model representations, we adopted a patch-based MIL architecture for the DR classification task. In the MIL setting, each image is treated as a bag $B$ containing multiple instances $x_i$, where each instance corresponds to an image patch:
\begin{equation}
  B = \{x_1, x_2, \ldots, x_n\},
  \label{eq:mil-bag}
\end{equation}

where $n$ represents the number of patches extracted from the image. Each instance is passed through a feature extractor $f(\cdot)$ to obtain a patch-level embedding $h_i$:
\begin{equation}
  h_i = f(x_i),
  \label{eq:patch-embedding}
\end{equation}
where $f(\cdot)$ represents the pretrained encoder. During feature extraction, positional embeddings are incorporated into the patch tokens. Depending on the encoder architecture, image-level patch representations were obtained using the appropriate pooling strategy (e.g. CLS token or average pooling). The complete pooling configurations are provided in Appendix C. To obtain an image-level representation, we apply attention-based MIL pooling, which aggregates patch-level features using learnable attention weights:
\begin{equation}
  z = \sum_{i=1}^{n} a_i h_i,
  \label{eq:mil-pooling}
\end{equation}
where $a_i$ denotes the attention weight assigned to the $i$-th patch. Following Ilse et al. [30], the attention weights are computed as:
\begin{equation}
  a_i = \frac{\exp\left(w^T \tanh\left(Vh_i^T\right)\right)}{\sum_j \exp\left(w^T \tanh\left(Vh_j^T\right)\right)}.
  \label{eq:attention-weight}
\end{equation}

The aggregated feature representation $z$ is then passed to a classifier to produce the final image-level prediction.

\subsubsection{Patch sampling during MIL inference and MIL training}
For the downstream task of disease classification, we took images from our external labelled training dataset, downsampled to 1024×1024 and split into a grid with 224×224 sized patches at a 12.5\% overlap, producing 25 instances per bag as shown in Fig. 1. Dense tiling with overlap at inference ensures comprehensive tissue coverage to ensure diagnostically relevant regions will be present in the bag. Each 224 × 224 image patch was processed independently by the pretrained encoder using the checkpoint’s internal token-level positional encoding. The resulting patch-level embedding was then augmented with a fixed two-dimensional sine-cosine embedding representing that patch’s location within the 5 × 5 UWF grid. These bag-level positional embeddings were added after encoder feature extraction and before MIL aggregation.

Bag-level classification was performed using an attention-based MIL pooling module that aggregates instance features, localises the most informative tiles and pools them together for the full image representation.  This attention module was made up of a linear layer, tanh activation, and a linear attention layer, and learns to assign weights to each patch representation and identify the most informative regions for prediction. The aggregated feature representation was passed through layer normalisation to stabilise feature scaling, and a linear classifier for final image-level prediction. Instance dropout was set to 0.3 to improve generalisation.

The feature encoder was frozen during MIL training so that the model evaluated the quality of the pretrained representations without further updating the encoder parameters. Only the attention-based MIL module and the final classifier layer were trained.

Training was performed using cross-entropy loss for the five-class DR task and binary cross-entropy with logits loss for the binary classification tasks. The Adam optimiser with a learning rate of $1 \times 10^{-3}$ was used, and a class-weighted loss to account for class imbalance. Downstream MIL training used a batch size of one bag per step, with 25 patches per bag. The models were trained for a maximum of 100 epochs using early stopping with a patience of 15 epochs based on validation AUROC for the binary classification tasks and validation Quadratic Weighted Kappa (QWK) for the five-class task. Because the encoder was frozen and only the lightweight MIL head was trained, all experiments were run on a single consumer GPU (NVIDIA GeForce RTX 3070).

\subsection{Evaluation of representation learning and downstream validation}
\subsubsection{High-Resolution Image Representation}
Conventional UWF image classification compresses the entire UWF into a single global representation following aggressive image resizing. In contrast, the proposed MIL framework aims to preserve local retinal information by representing each image as a collection of patches. Here, we first investigated how best to utilise high-resolution UWF images by comparing four image representation strategies using the same frozen encoder (DINOv3):
\begin{itemize}
    \item \textbf{Global:} The complete UWF image was resized to the native input resolution of the encoder (224×224) and represented by a single image-level embedding (mean of token embeddings). This embedding was passed through layer normalisation followed by a linear classifier.
    \item \textbf{Patch-based:} Each complete UWF image was resized to 1024×1024 and divided into a regular 5×5 grid of overlapping 224×224 patches, as detailed in Section III.B.2. Each patch was independently encoded by the same frozen DINOv3 backbone and the resulting patch embeddings were aggregated using one of three strategies:
    \begin{itemize}
        \item Mean pooling
        \item Max pooling
        \item Attention-based pooling
    \end{itemize}
\end{itemize}
For all four strategies, the same encoder weights, learning rate, downstream classifier, optimiser, and training protocol were used. To enable a fair comparison against the Global baseline and to keep this preliminary comparison computationally efficient, instance dropout was disabled and models were trained with a reduced early-stopping patience (patience = 5). Consequently, results in this section are not directly comparable to the main encoder comparison in Section IV.A.2, which used the full training protocol described in Section III.B.2.

\subsubsection{Pretrained Feature Extractors within the MIL framework}
To evaluate the suitability of different backbone architectures for patch-based MIL, we compared CNN-based and ViT-based feature extractors within the same MIL framework. In all experiments, the MIL aggregation module and training procedure were kept identical, while only the feature extractor was varied. Feature extractors were frozen and only the MIL module and classifier trained. 
We evaluated the following models:

\textit{CNN baselines:}
\begin{itemize}
    \item \textbf{ResNet+MIL:} ResNet-18 or ResNet-50 [31] pretrained on ImageNet-1k [20].
\end{itemize}

\textit{Controlled ImageNet-pretrained ViT-B comparison:}
\begin{itemize}
    \item \textbf{ViT-MAE(ImageNet)+MIL:} ViT-B pretrained on ImageNet-1k using Masked Autoencoder [5].
    \item \textbf{ViT-Supervised(ImageNet)+MIL:} ViT-B pretrained on ImageNet-1k using classification-based supervised learning [4].
    \item \textbf{ViT-DINOv1(ImageNet)+MIL:} ViT-B pretrained on ImageNet-1k using DINOv1 self-distillation [6].
\end{itemize}

\textit{Additional pretrained encoders:}
\begin{itemize}
    \item \textbf{ViT-MAE(AlzEye)+MIL:} A domain-specific ViT-S encoder pretrained using Masked Autoencoder on approximately 100,000 unlabelled UWF Optomap retinal images. Details of the pretraining procedure are provided in Appendix A.
    \item \textbf{ViT-DINOv3(LVD)+MIL:} ViT-B pretrained on LVD-1689M using DINOv3 self-distillation [7].
\end{itemize}

The domain-specific MAE encoder used a ViT-S architecture and was pretrained on approximately 100,000 unlabelled Optomap images, whereas the publicly available MAE(ImageNet) encoder used a larger ViT-B architecture pretrained on over 1 million natural images. Consequently, comparisons between these two MAE models should be interpreted cautiously, as they differ in encoder architecture, model capacity, and pretraining data scale. RETFound [8] was not included as an additional domain-specific encoder because it was pretrained on CFP and OCT rather than UWF Optomap images, and therefore is not modality-matched to the target domain, and because its ViT-L architecture would introduce further mismatch. Details of the domain-specific pretraining procedure are provided in Appendix A. All other models used a ViT-base architecture to avoid differences caused by model capacity and isolate the effect of the pretraining strategy. The ImageNet-pretrained models were selected to enable comparisons under a common pretraining dataset. In addition, we included both DINOv1 and DINOv3 to examine whether recent advances in self-distillation and large-scale pretraining translate to improved performance in a weakly supervised ophthalmic imaging setting. ResNet-50 was used for the five-class DR task and ResNet-18 for the binary classification tasks, matching the CNN capacity to dataset scale.

Accordingly, the controlled comparison of pretraining objectives was based primarily on the supervised, MAE, and DINOv1 ViT-B encoders pretrained on ImageNet. DINOv3 was additionally evaluated as a contemporary large-scale self-distillation model, while the domain-specific MAE was included as a complementary comparison and was not used to isolate the effect of pretraining objective.

\subsubsection{Representation Analysis}
To better understand how different feature extractors interact with the MIL module for classification tasks, we generated heatmap visualisations of the learned attention weights assigned to individual image patches. For each image, attention weights produced by the MIL pooling module were extracted after the SoftMax operation and min-max normalised to the range [0,1]. The resulting values were projected back onto the corresponding patch locations to produce patch attention heatmaps, where higher intensities indicate a greater contribution to the final classification decision. These attention weights were interpreted as properties of the learned aggregation mechanism and were not treated as validated lesion-level explanations or segmentation maps. Qualitative and quantitative comparisons were performed across feature extractors using representative examples from the test set. Quantitative attention statistics were calculated from the unnormalised SoftMax attention weights before min-max visualisation.

In addition, we visualised bag-level feature representations using Uniform Manifold Approximation and Projection (UMAP). For each image, the aggregated bag representation produced by the MIL module prior to the final classification layer was extracted and projected into a two-dimensional space to enable qualitative comparison of feature space organisation across the different pretraining strategies.

\subsubsection{Partial Encoder Fine-tuning}
Although frozen feature extractors provide a controlled setting for comparing pretrained representations, they may not fully exploit the downstream task, particularly when the pretraining objective differs substantially from the target application. To investigate whether weaker frozen representations could benefit from limited task-specific adaptation, we additionally evaluated a partially fine-tuned setting in which only the final encoder block was unfrozen, while all preceding layers remained frozen. The unfrozen encoder parameters were optimised using a lower learning rate of $5 \times 10^{-5}$, while the MIL aggregation module and classifier retained the learning rate of $1 \times 10^{-3}$. The MIL aggregation module, classifier, and training protocol were otherwise unchanged, allowing the effect of limited encoder adaptation on downstream performance to be assessed. Each partially fine-tuned experiment was repeated using the same three seeds and fixed data splits as the corresponding frozen-encoder experiment.

\subsubsection{Downstream Evaluation Metrics}
The proposed framework was evaluated on three downstream classification datasets, each serving a complementary role within the experimental design. The MMRDR dataset was used as the primary analysis dataset, while the DeepDRiD and intraocular tumour datasets were used to assess external and cross-disease evaluation, respectively.

Evaluation protocols differed slightly depending on the dataset. For the MMRDR and DeepDRiD datasets, performance was evaluated on the unseen test sets. For the intraocular tumour dataset, performance was evaluated using 5-fold stratified cross-validation across folds. 

For the five-class DR grading task, we reported the following metrics:
\begin{itemize}
    \item Quadratic Weighted Kappa (QWK)
    \item Area Under the Receiver Operating Characteristic Curve (AUROC)
\end{itemize}

QWK measures agreement between predicted and true labels while accounting for the ordinal nature of DR severity. We used this as the primary metric due to its suitability for grading tasks [32], as misclassifications between distant classes are penalised more than those between adjacent classes. Multi-class AUROC was computed using a one-vs-rest approach and averaged across classes using class-support weighting.

For the binary classification tasks, performance was primarily evaluated using AUROC. Additional metrics, including weighted F1-score, accuracy, area under the precision--recall curve (AUPR), and Cohen's $\kappa$, were also computed for all experiments and are reported in Appendix D. Accuracy and Cohen’s $\kappa$ were calculated using a decision threshold selected on the validation set to maximise classification accuracy; the selected threshold was then applied unchanged to the test set.

\section{Results}
\subsection{Validation}
\subsubsection{Representation Strategy}
We compared four representation strategies using the same frozen encoder (DINOv3). Under this experimental setting, representing UWF images using multiple higher-resolution local patches generally improved performance relative to representing the complete image using a single low-resolution embedding, as shown in Table I. Dividing the full image into patches and averaging patch-level representations increased the QWK (from 0.7999 to 0.8358) and AUROC (from 0.8423 to 0.8799), demonstrating that preserving local retinal information can provide more discriminative image representations than compressing the entire retina into a single global embedding. Max pooling produced the weakest performance (QWK = 0.7895), suggesting that disease severity cannot be reliably characterised by the most salient retinal region alone. In contrast, the proposed attention-based MIL framework achieved the highest performance (QWK = 0.8621), suggesting that adaptive aggregation of multiple retinal regions provides the most informative image representation for DR grading. Results are based on a single run, with a reduced protocol.

\begin{table}[!t]
\caption{Comparison of representation strategies for UWF images using the same frozen encoder (DINOv3). Best results are shown in bold.}
\label{tab:representation-strategies}
\centering
\begin{tabular}{lcc}
\hline
\textbf{Representation strategy} & \textbf{QWK} & \textbf{AUROC} \\
\hline
Global & 0.7999 & 0.8423 \\
Patch--mean & 0.8358 & 0.8799 \\
Patch--max & 0.7895 & 0.8439 \\
Patch--attention & \textbf{0.8621} & \textbf{0.8922} \\
\hline
\end{tabular}
\end{table}

\subsubsection{Multiple-Instance Learning Framework}
To evaluate the suitability of different backbone architectures for patch-based MIL, we applied the same attention-based MIL framework while varying the feature extractor across three UWF datasets: multi-class DR grading (MMRDR [18]), binary DR classification (DeepDRiD [27]), and intraocular tumour classification [28]. Results are shown in Table II.

Across the three datasets, supervised and self-distillation-based Vision Transformers generally achieved the strongest performance, although the relative ranking of the reconstruction-based models varied across tasks. Within the controlled ImageNet-pretrained ViT-B comparison, supervised and self-distillation-based pretraining outperformed reconstruction-based MAE on the primary MMRDR benchmark, indicating superior transferability of frozen representations within the proposed MIL framework. DINOv3 achieved the highest mean QWK (0.8629 ± 0.0102), marginally above DINOv1 (0.8602 ± 0.0034); this difference lies within one standard deviation and the two self-distillation encoders should be regarded as performing comparably on this benchmark.

In contrast, the ranking shifted on the smaller binary DeepDRiD task. DINOv3 remained the strongest model (AUROC = 0.9030), but ViT-MAE(ImageNet) achieved an AUROC of 0.8131, comparable to DINOv1 (0.8136), suggesting that the relative advantage of discriminative pretraining was less pronounced for binary DR classification than for five-class grading. The larger standard deviations for several models on DeepDRiD also reflect greater variation across runs on this small dataset.

The intraocular tumour task provided a further test of transferability beyond DR classification. The supervised and DINO-based encoders achieved the strongest performance, with DINOv1 reaching an AUROC of 0.9963. 

On DeepDRiD, ViT-MAE(AlzEye)+MIL did not perform above chance (AUROC = 0.5207 ± 0.0353). On the five-class MMRDR task the same encoder performed above chance (QWK = 0.3247 ± 0.0474), but substantially below all general-purpose encoders. The comparatively weak performance of ViT-MAE(AlzEye) on both DR datasets suggests that the learned retinal-domain representations transferred less effectively to DR classification than the large-scale general-purpose discriminative encoders when used as frozen feature extractors within the MIL framework. In contrast, its comparatively stronger performance on intraocular tumour classification indicates that these representations retained information useful for distinguishing more pronounced abnormalities. However, this observation should be interpreted cautiously because the domain-specific model differs from the general-purpose encoders in architecture, model capacity, pretraining scale, and training configuration.

Further analysis of the performance of the general-purpose pretrained encoders in the five-class DR task showed that ViT-MAE(ImageNet) struggled to separate adjacent DR severity grades (see Fig. 2). Amongst the higher-performing ViT-based models, confusion matrices revealed a common challenge in distinguishing between mild and moderate DR, highlighting the subtle visual differences between these two stages. While all models struggled to differentiate between these adjacent grades, DINOv3 showed improved recognition of mild DR relative to the others. The supervised ViT model achieved the strongest performance on separating healthy images from DR cases at all stages, whereas DINOv1 demonstrated the strongest separation of severe DR cases.

\begin{table*}[!t]
\caption{Performance comparison across the three downstream UWF classification tasks. Dashes indicate that a model was not evaluated on that dataset. Values for MMRDR and DeepDRiD are mean $\pm$ standard deviation across three training seeds. Values for the tumour dataset are mean $\pm$ standard deviation across the five held-out cross-validation folds. Best results are shown in bold.}
\label{tab:main_results}
\centering
\scriptsize
\setlength{\tabcolsep}{4pt}
\renewcommand{\arraystretch}{1.15}
\begin{tabular}{@{}lcccc@{}}
\toprule
\textbf{Model} &
\multicolumn{2}{c}{\textbf{Five-class DR (MMRDR)}} &
\textbf{Binary DR} &
\textbf{Binary Tumour} \\
\cmidrule(lr){2-3}
&
\textbf{QWK} &
\textbf{AUROC} &
\textbf{AUROC} &
\textbf{AUROC} \\
\midrule

\multicolumn{5}{l}{\textit{Vision Transformers}}\\

ViT-MAE (ImageNet)+MIL &
$0.5325 \pm 0.0205$ &
$0.7282 \pm 0.0097$ &
$0.8131 \pm 0.0053$ &
$0.8742 \pm 0.0199$ \\

ViT-Supervised (ImageNet)+MIL &
$0.8396 \pm 0.0043$ &
$0.8781 \pm 0.0026$ &
$0.8520 \pm 0.0442$ &
$0.9806 \pm 0.0092$ \\

ViT-DINOv1 (ImageNet)+MIL &
$0.8602 \pm 0.0034$ &
$\mathbf{0.8963 \pm 0.0017}$ &
$0.8136 \pm 0.0616$ &
$\mathbf{0.9963 \pm 0.0011}$ \\

ViT-DINOv3 (LVD)+MIL &
$\mathbf{0.8629 \pm 0.0102}$ &
$0.8958 \pm 0.0051$ &
$\mathbf{0.9030 \pm 0.0244}$ &
$0.9894 \pm 0.0037$ \\

\midrule

\multicolumn{5}{l}{\textit{CNN baselines}}\\

ResNet-50 (ImageNet)+MIL &
$0.6907 \pm 0.0016$ &
$0.7804 \pm 0.0049$ &
-- &
-- \\

ResNet-18 (ImageNet)+MIL &
-- &
-- &
$0.6520 \pm 0.1576$ &
$0.8478 \pm 0.1370$ \\

\midrule

\multicolumn{5}{l}{\textit{Domain-specific ViT-S}}\\

ViT-MAE (AlzEye)+MIL &
$0.3247 \pm 0.0474$ &
$0.6301 \pm 0.0034$ &
$0.5207 \pm 0.0353$ &
$0.8613 \pm 0.0101$ \\

\bottomrule
\end{tabular}
\end{table*}

\begin{figure}[!t]
\centering
\includegraphics[width=\columnwidth]{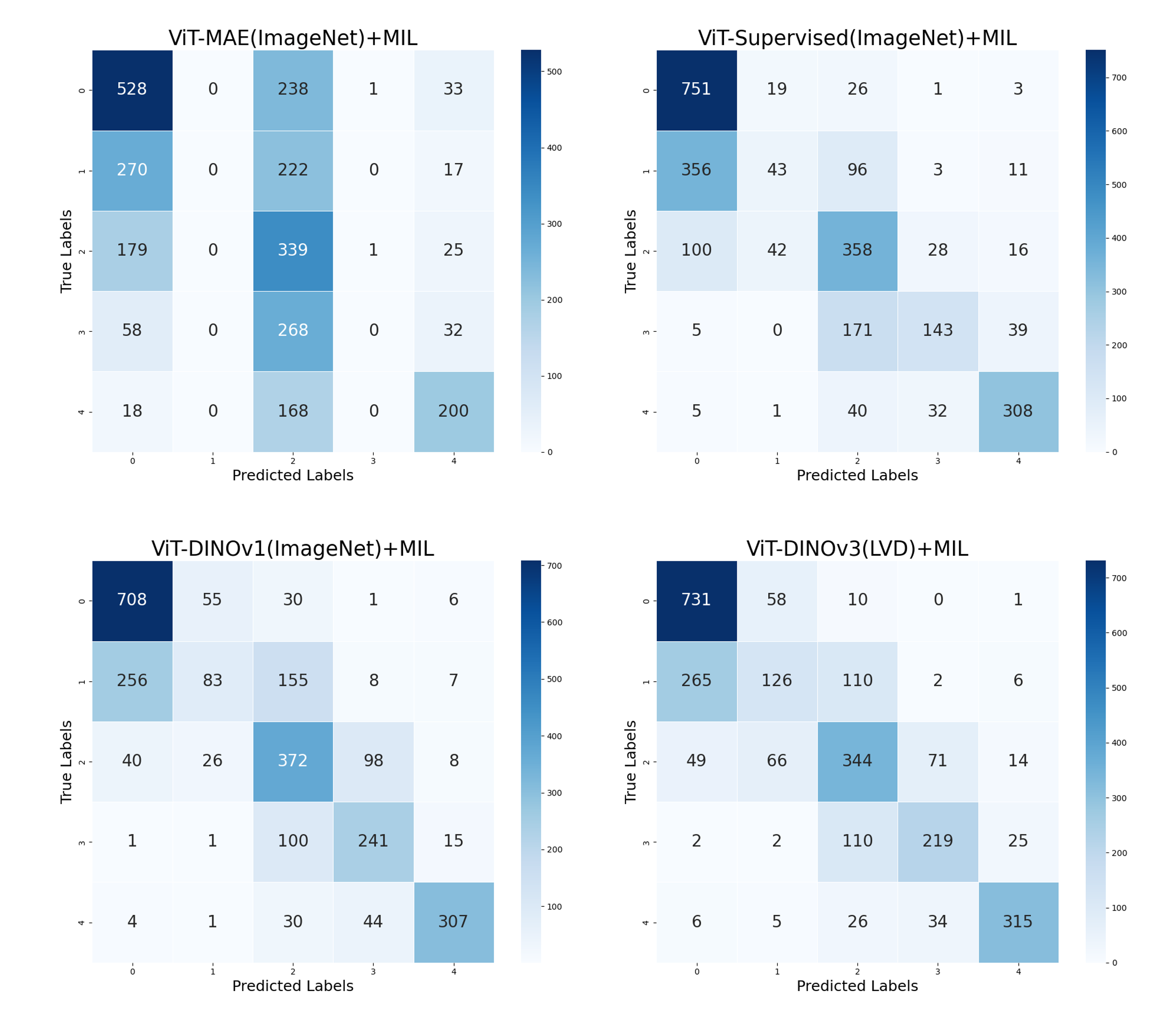}
\caption{Confusion matrices showing the performance of selected encoder+MIL models on the unseen test set. Labels denote 0: no DR, 1: mild DR, 2: moderate DR, 3: severe DR, and 4: proliferative DR (pDR).}
\label{fig:confusion-matrices}
\end{figure}

To examine the learned representations in feature space, we visualised bag-level embeddings using UMAP (Fig. 3). ViT-MAE showed a highly intermixed feature space with no clear visual separation between the disease grades. In contrast, the supervised and DINO-based models exhibited more structured organisations, with pDR samples tending to occupy distinct regions of the embedding space. However, substantial overlap between adjacent grades remained, particularly between no DR, mild, and moderate DR, consistent with the confusion matrix analysis.

\begin{figure}[!t]
\centering
\includegraphics[width=\columnwidth]{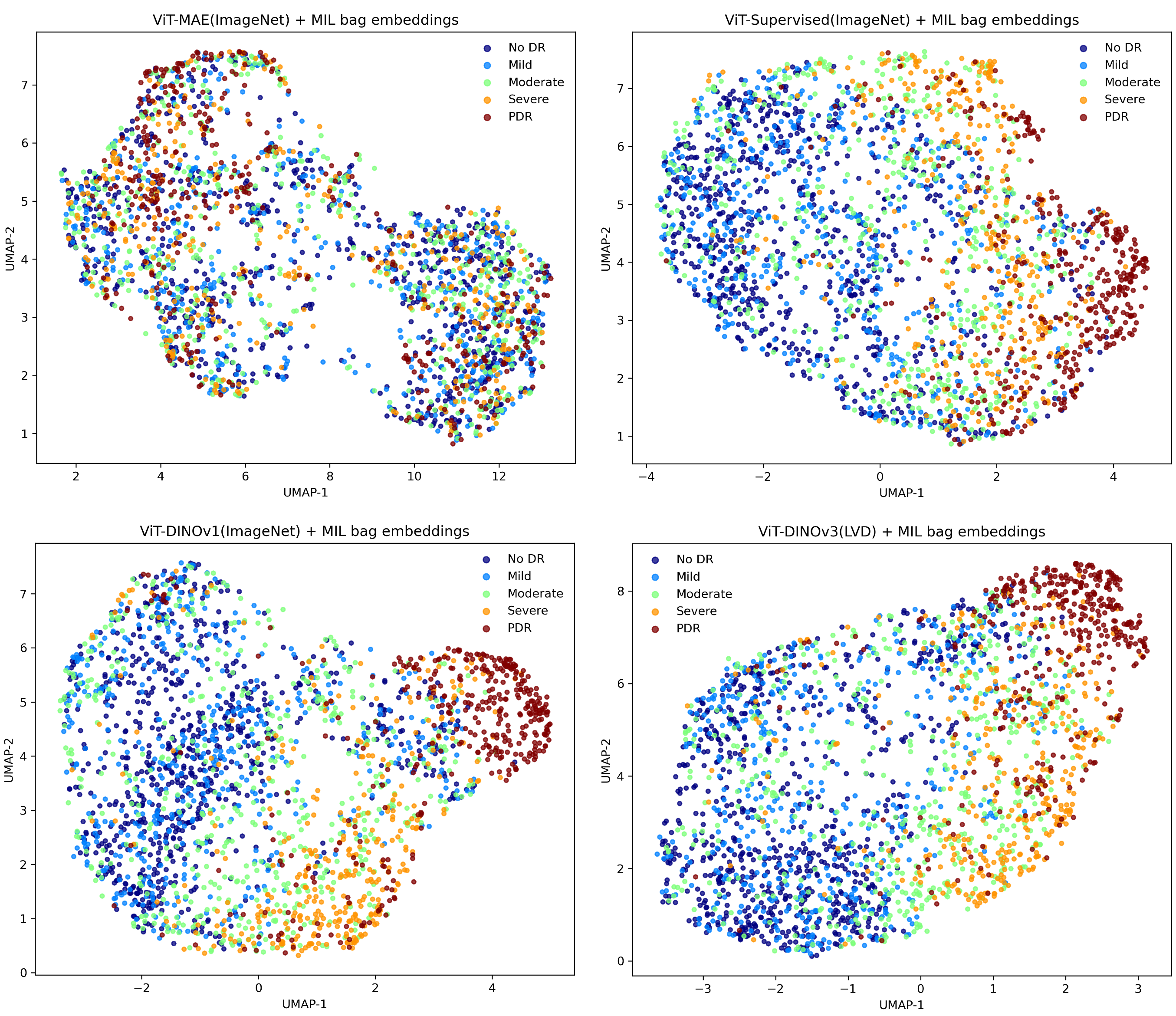}
\caption{UMAP visualisations of bag-level embeddings extracted from the MIL framework for each pretrained encoder. Each point represents a test image and is coloured according to DR severity grade.}
\label{fig:umap-embeddings}
\end{figure}

\begin{figure*}[!t]
\centering
\includegraphics[width=\textwidth]{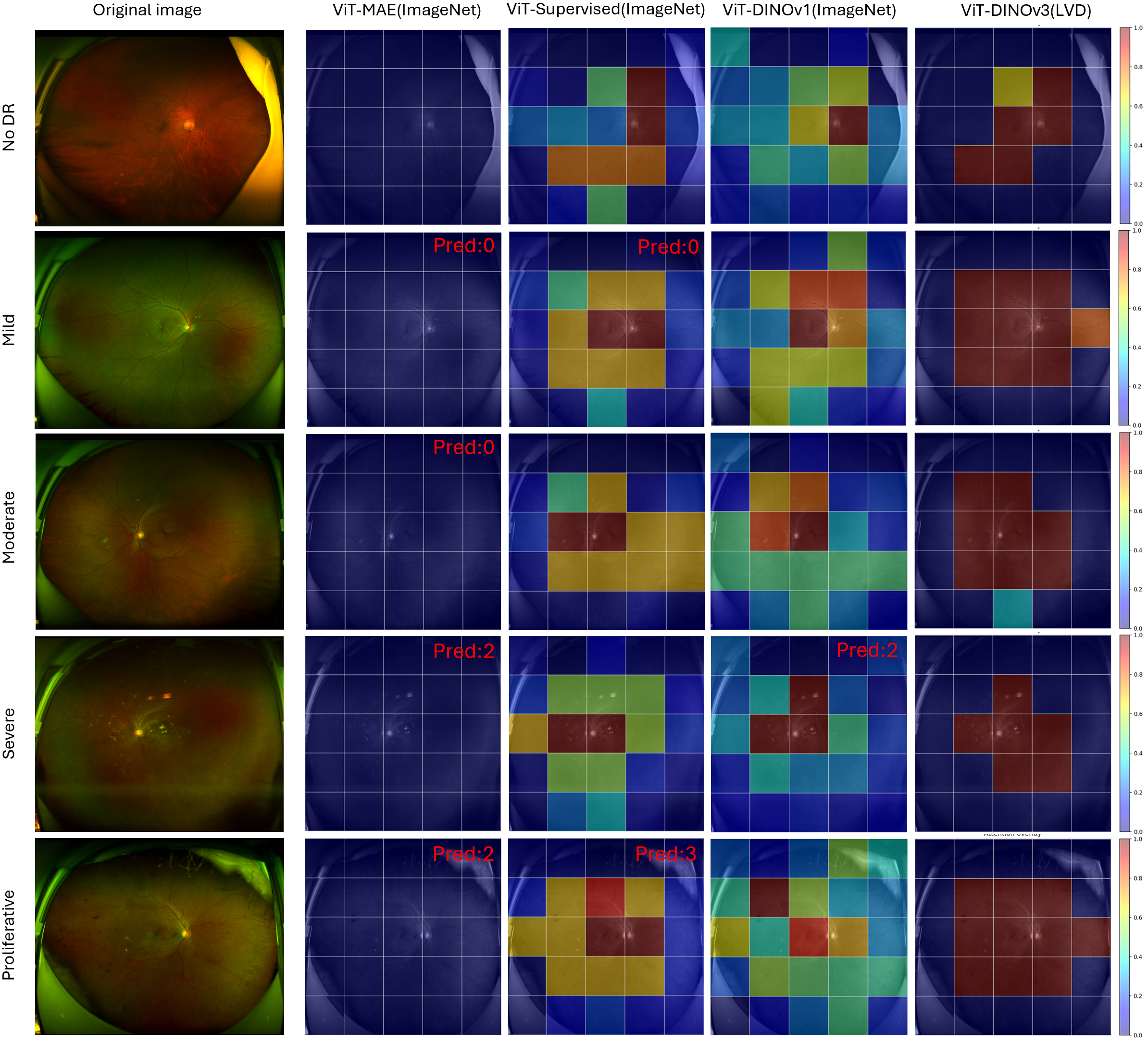}
\caption{Attention heatmaps for one example from each class in the MMRDR dataset. Incorrect classifications are labelled in red with their predicted classes.}
\label{fig:attention-heatmaps}
\end{figure*}

\subsection{Visualisations of Model Attention}
To qualitatively assess how different feature extractors utilise retinal information for downstream tasks, we visualised the distribution of attention to image patches for representative examples. These visualisations were generated by projecting the MIL attention weights onto the corresponding patch locations and overlaying them on the reconstructed patch mosaic. Attention weights were min-max normalised and heatmaps were generated with the same scale. We observed notable differences between models.  

\subsubsection{MMRDR Attention Analysis}
Attention heatmaps were generated for the MMRDR test set to visualise the distribution of attention by the MIL module. Fig. 4 shows a representation of attention heatmaps for different encoder+MIL combinations across each DR severity grade, with one test image selected from each severity grade to illustrate the recurrent attention patterns observed across the dataset. Clear differences were observed in how each model distributed attention across the retina. 

The ViT-MAE(ImageNet) encoder consistently produced uniform attention maps with no distinction between retinal regions. This behaviour was accompanied by multiple incorrect predictions, particularly for pathological cases, suggesting that the MIL module was unable to identify discriminative patch-level representations from the frozen MAE features. This was confirmed quantitatively by a near-zero standard deviation of attention weights (Fig. 5), indicating that the model assigned equal importance to all image patches. Consequently, the attention-based aggregation effectively behaved similarly to an unweighted averaging of patch representations, providing little advantage over simple mean aggregation.

In contrast, the supervised ViT and ViT-DINOv1 encoders produced more structured attention maps, assigning greater importance to central retinal regions. The supervised ViT typically concentrated attention on the optic nerve head. ViT-DINOv1 exhibited a more graded attention distribution, characterised by one or two dominant patches surrounded by neighbouring patches with progressively lower attention weights. In comparison, the ViT-DINOv3 encoder assigned relatively high attention across a broader region of the posterior retina rather than concentrating on single dominant patches. Attention remained focused on the retinal area while largely avoiding the peripheral image borders and non-retinal background.

These observations were supported by quantitative analysis (Fig. 5). Compared with ViT-MAE(ImageNet), the supervised ViT and both ViT-DINO models allocated substantially greater cumulative attention to their three highest-weighted patches, confirming that the MIL module learned to prioritise specific retinal regions. The supervised ViT exhibited the greatest attention concentration, whereas ViT-DINOv1 showed a more gradual spatial weighting. DINOv3 exhibited the highest per-image attention variability (Fig. 5b) yet the lowest top-3 attention mass among the discriminative encoders (Fig. 5a). This indicates that it separated central retinal patches from the low-attention periphery while distributing its highest weights across several central patches rather than a small number of dominant instances.

\begin{figure}[!t]
\centering
\includegraphics[width=\columnwidth]{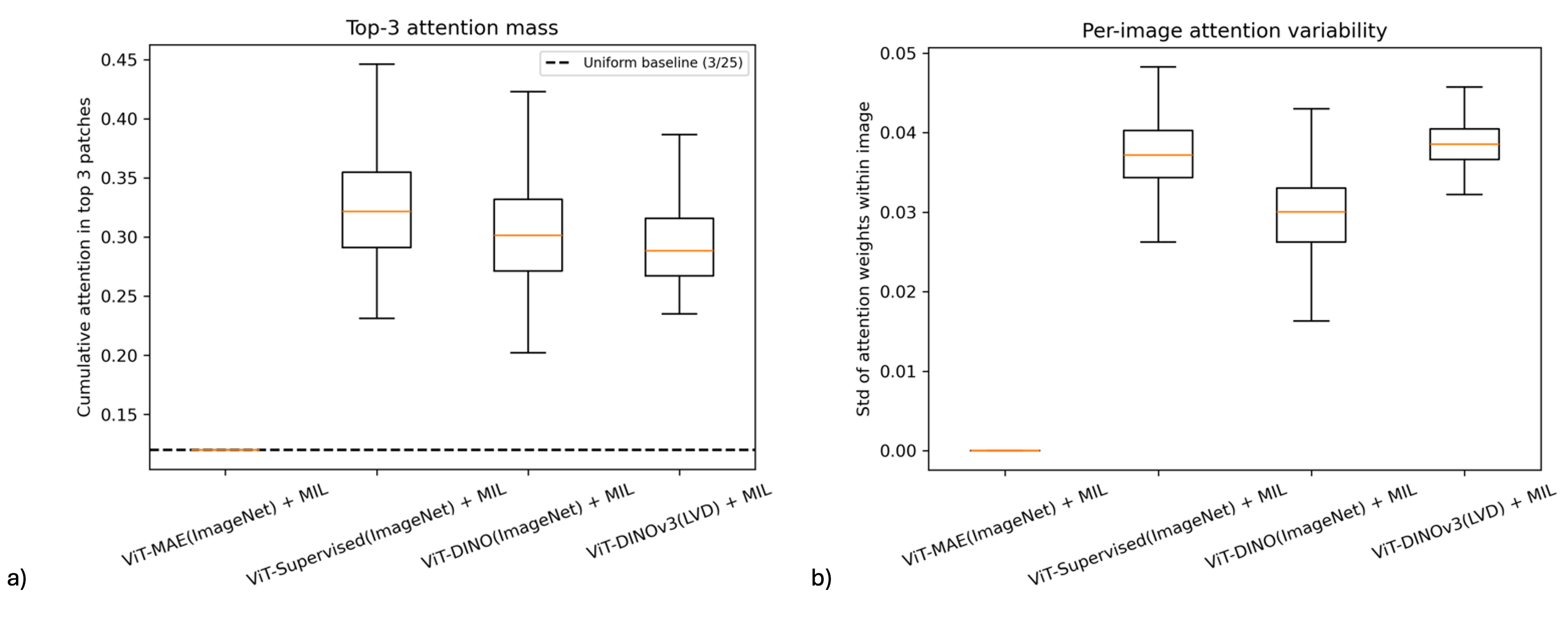}
\caption{Attention-weight distributions for each encoder+MIL model: (a) cumulative attention assigned to the three highest-weighted patches in each image and (b) variability of attention weights across patches within each image.}
\label{fig:attention-distributions}
\end{figure}

\subsubsection{Cross-disease Analysis}
Fig. 6 presents representative attention maps for two intraocular tumour cases using the different pretrained feature extractors. Unlike the primary MMRDR analysis in Fig. 4, this figure includes the ResNet baseline and domain-specific AlzEye MAE to compare attention behaviour across both encoder architectures and pretraining domains. DINOv3 was omitted to maintain figure readability, as its attention behaviour was already characterised in the primary analysis. Again, the ViT-MAE(ImageNet) model exhibited uniform attention distribution across all patches, with no distinct patches receiving higher attention weighting. Interestingly, ViT-MAE(AlzEye) produced more structured attention than ViT-MAE(ImageNet), suggesting that domain-specific pretraining learned discriminative features, even if this did not translate into the best classification performance. 

The remaining encoder+MIL models demonstrated higher variability between attention across patches. The other ViT-based models and ResNet-18 baseline concentrated attention in specific regions, typically focusing on a small subset of high-weighted patches within the central retinal region, as observed qualitatively in the heatmaps. For images containing tumours, these models assigned higher aggregation weights to spatially localised retinal regions, some of which visually coincided with the tumour region in the illustrated examples. In healthy images, attention was frequently centred around the optic nerve head (Fig. 6b). 

Notably, certain feature extractors appeared to cause the MIL module to adopt different strategies to identify relevant features. For the supervised feature extractors, attention was focused on fewer, highly weighted patches. For the self-supervised feature extractors, attention was distributed across a broader area.

\begin{figure*}[!t]
\centering
\includegraphics[width=\textwidth]{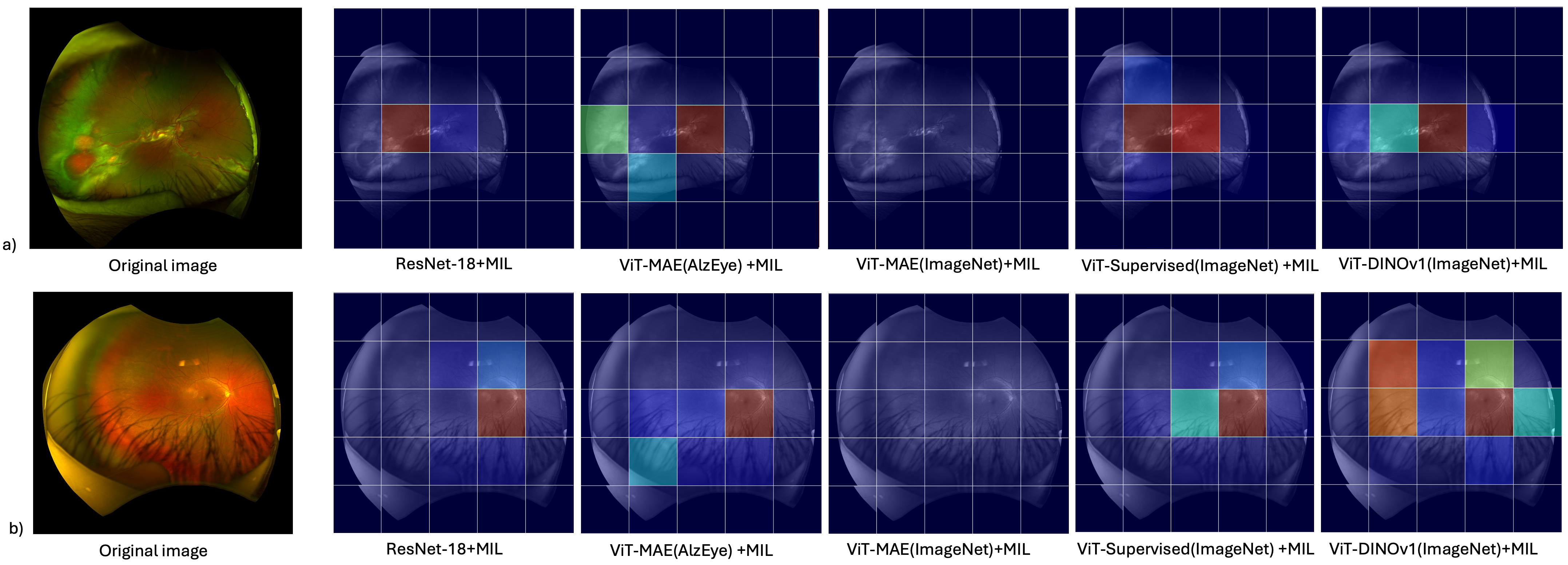}
\caption{Attention heatmaps for selected encoder+MIL models using two examples from the evaluation set: (a) a retinal image with an intraocular tumour and (b) a retinal image without a tumour.}
\label{fig:tumour-attention-heatmaps}
\end{figure*}

\subsubsection{Generalisation of Attention Behaviour}
In the non-referable/referable DR classification task, similar trends were observed to those seen in the intraocular tumours dataset for models using supervised and distillation-based feature extractors. ViT-Supervised(ImageNet), ViT-MAE(ImageNet), ViT-DINOv1(ImageNet), ViT-DINOv3(LVD), and ResNet-18(ImageNet) demonstrated consistent attention patterns.

However, models based on the domain-specific Masked Autoencoder exhibited less reliable attention behaviour. Examples can be seen in Appendix Fig. 7. The ViT-MAE(AlzEye)+MIL model showed unexpected behaviour, where attention was assigned to peripheral regions, particularly on non-retinal regions. This suggests the downstream MIL module identified and relied on non-retinal artefacts, and this sensitivity resulted in a worse classification performance. 

\subsection{Further Adaptation of Encoders}
To investigate whether weaker frozen representations could benefit from limited task-specific adaptation, we repeated the MMRDR experiments with only the final encoder block of each Vision Transformer unfrozen. Table III compares the QWK scores obtained using frozen and partially fine-tuned encoders across three runs.

Partial fine-tuning improved performance for all evaluated Vision Transformer models, although the magnitude of improvement differed substantially between pretraining strategies. The largest improvements were observed for the MAE-based encoders (+0.2875 QWK for ViT-MAE(ImageNet) and +0.2192 QWK for ViT-MAE(AlzEye)), whereas the supervised ViT and DINO-based models exhibited comparatively smaller gains. This finding suggests that the frozen discriminative representations learned through supervised and self-distillation objectives transfer more effectively to the proposed patch-based MIL framework, while reconstruction-based representations benefit more substantially from downstream adaptation.

Despite the substantial performance improvement observed for the MAE-based encoders, no obvious qualitative changes were observed in the learned attention distributions across representative test images. The primary difference was an increase in prediction confidence rather than a change in the spatial weighting of retinal regions. These observations suggest that partial fine-tuning primarily improved the encoded patch representations rather than altering how the MIL module aggregated spatial information.

\begin{table*}[!t]
\caption{Comparison of QWK scores for five-class DR classification on MMRDR using frozen and partially fine-tuned feature extractors. Partial fine-tuning unfreezes the final encoder block. Values are reported as mean $\pm$ standard deviation across three runs. Improvement denotes the absolute difference between the mean partially fine-tuned and frozen QWK values.}
\label{tab:partial-finetuning}
\centering
\begin{tabular}{lccc}
\hline
\textbf{Model} & \textbf{Frozen QWK} & \textbf{Partially fine-tuned QWK} & \textbf{Improvement} \\
\hline
ViT-MAE (AlzEye)+MIL & $0.3247 \pm 0.0474$ & $0.5439 \pm 0.0199$ & $+0.2192$ \\
ViT-MAE (ImageNet)+MIL & $0.5325 \pm 0.0205$ & $0.8200 \pm 0.0144$ & $+0.2875$ \\
ViT-Supervised (ImageNet)+MIL & $0.8396 \pm 0.0043$ & $0.8691 \pm 0.0078$ & $+0.0301$ \\
ViT-DINOv1 (ImageNet)+MIL & $0.8602 \pm 0.0034$ & $0.8843 \pm 0.0047$ & $+0.0241$ \\
ViT-DINOv3 (LVD)+MIL & $0.8629 \pm 0.0102$ & $0.8947 \pm 0.0025$ & $+0.0318$ \\
\hline
\end{tabular}
\end{table*}

\section{Discussion}
In this study, we investigated how foundation model representations transfer to high-resolution UWF retinal image analysis within a controlled patch-based MIL framework. By combining patch-level feature extraction with attention-based aggregation, the proposed framework enabled different pretrained representations to be compared while preserving the spatial information present in UWF images. Beyond comparing pretraining strategies, the framework also allowed us to investigate how high-resolution image representation and limited encoder adaptation influence downstream performance.	

UWF retinal images frequently contain pathological biomarkers distributed across multiple retinal regions. Unlike conventional image classification, which compresses the entire retina into a single representation following aggressive image downsampling, the proposed patch-based framework preserves local retinal detail throughout feature extraction. The improved performance of patch-based representations with mean or attention pooling suggests that preserving these local features benefits DR grading, although requiring more encoder evaluations than the global baseline. Interestingly, selecting only the strongest patch response through max pooling resulted in the lowest performance, suggesting that DR severity is not determined by a single dominant feature, but rather by the cumulative distribution of pathological features across the retina. Attention-based MIL further improved performance by learning to adaptively weight information from the multiple retinal regions. This is consistent with clinical assessment of DR images, where disease grading is based on the combined distribution and severity of lesions, including haemorrhages, microaneurysms, venous abnormalities, and neovascularisation. This may be particularly relevant for UWF imaging, where clinically important lesions might extend into the retinal periphery and compressing the entire field of view inevitably reduces the spatial resolution available to identify these peripheral abnormalities.

Within this framework, our experiments demonstrated that discriminative Vision Transformers were particularly well suited as feature extractors within the MIL framework. Among the ViT-based models, the supervised and DINO-based encoders consistently achieved the strongest downstream performance. Although the supervised and DINO-based models achieved similar overall performance, each demonstrated strengths in recognising different disease stages. The supervised ViT model achieved the strongest performance on separating any DR cases from healthy images, the DINOv1-based model demonstrated improved separation of severe DR cases, and the DINOv3-based model improved recognition of mild DR. These findings suggest that different pretraining strategies learn distinct retinal characteristics, which subsequently influence downstream classification performance within the MIL framework.	The original MMRDR benchmarks reported higher performance for some fully fine-tuned models (ViT-Supervised Accuracy = 0.69) than that achieved by the frozen-encoder configurations in this study [18]. This difference is expected because the present work was designed to assess the transferability of frozen pretrained representations rather than maximise classification performance. Consistent with this distinction, partially unfreezing the final encoder block improved performance across all evaluated ViTs.

Attention analysis provided insight into the differing behaviour of the pretrained representations interacting with the MIL module.  For example, when analysing a healthy image, the supervised ViT-based model tended to concentrate attention on a smaller number of patches, primarily around the optic nerve head, whereas the DINO-based model distributed attention over a larger region of the retina. Quantitative analysis also supported the difference in attention distribution, with the DINOv3-based model demonstrating structured attention while avoiding peripheral image borders and non-retinal background. Interestingly, attention was not assigned to all visible pathological features. Instead, the models consistently prioritised a subset of retinal regions while assigning relatively low attention to other abnormalities. This behaviour is expected within an MIL framework, where the attention mechanism learns which patch representations are most informative for image-level classification rather than performing exhaustive lesion localisation.

In contrast, the ViT-MAE(ImageNet) encoder exhibited different attention behaviour across the downstream tasks. For the five-class DR grading and intraocular tumour datasets, attention remained uniform, indicating little distinction between retinal regions. These findings suggest that the frozen MAE representations did not provide sufficiently discriminative local features for the MIL module to reliably identify potentially informative retinal regions. Consequently, although these representations remained adequate for the simpler binary classification tasks, they struggled to distinguish the more subtle local differences required for multi-class DR grading. 

After unfreezing the final encoder block, both ViT-MAE(AlzEye) and ViT-MAE(ImageNet) demonstrated substantial improvements in downstream performance (+0.2192 and +0.2875 QWK, respectively; see Table III), supporting the view that MAE representations benefit substantially from downstream adaptation. Notably, after partial fine-tuning the MAE(ImageNet) encoder reached QWK = 0.8200 ± 0.0144, comparable to the frozen supervised ViT (QWK = 0.8396 ± 0.0043). These findings suggest that the observed disadvantage of MAE is largely confined to the frozen transfer setting investigated in this study. Consequently, reconstruction-based encoders may remain competitive when downstream adaptation is feasible, whereas the choice of pretraining objective appears most influential when pretrained representations are used as fixed feature extractors.

Within the controlled comparison of ViT-B encoders pretrained on ImageNet, supervised and self-distillation-based pretraining produced substantially more transferable frozen representations than reconstruction-based MAE pretraining. This may reflect the nature of the training objectives, where the Masked Autoencoder focuses on reconstruction and may not prioritise class-relevant features. Self-distillation-based pretraining (DINO) appears to support more selective attention patterns, ultimately resulting in improved discrimination between DR severity grades. This observation suggests that the effectiveness of MIL in medical imaging analysis depends not only on the aggregation strategy itself, but also on the transferability and clinical relevance of the underlying feature representations.

\section{Limitations}
This study has several limitations. First, qualitative analysis of the attention maps suggested that certain anatomical structures, particularly the optic nerve head, frequently received high attention weights. While this may reflect clinically relevant contextual information, it also highlights a broader limitation of weakly supervised approaches, including MIL, where models may exploit consistent but non-causal patterns in the data. Other examples include motion or peripheral artefacts being more prevalent in pathological cases due to patient discomfort, or device and operator-specific acquisition biases. Such patterns may be inadvertently associated with disease labels and subsequently exploited by the MIL framework. Further investigation is therefore required to disentangle true pathological signals from acquisition-related biases, for example through controlled datasets, experiments designed to isolate acquisition conditions, and bias mitigation strategies based on standardised preprocessing.

The potential benefits of patch-based learning may be underestimated in this study. The UWF images used for training and evaluation were resized to 1024 × 1024 pixels for computational feasibility, rather than using the native Optomap resolution of approximately 4000 × 4000 pixels. While this resolution preserves substantially more spatial information than conventional image classification pipelines, it does not fully capture the potential advantages of patch-based learning on true full-resolution UWF images.

Evaluation was limited by the small number of publicly available labelled UWF retinal datasets. Compared with conventional CFP, labelled UWF datasets remain relatively scarce, and the available datasets are typically small, class-imbalanced, and have limited evaluation protocols. Although the datasets used in this study span both DR grading and intraocular tumour classification, it remains unclear whether the observed differences between pretraining strategies generalise to other UWF cohorts, imaging technologies, grading protocols, or retinal diseases. Furthermore, patient identifiers were unavailable for the intraocular tumour dataset, preventing verification of subject-level separation between training and evaluation partitions. Consequently, the possibility of the same eye appearing across training and evaluation folds cannot be excluded, and the performance on this dataset should be interpreted with appropriate caution.

A common preprocessing pipeline was applied across all encoders rather than checkpoint-specific normalisation. Although this ensured a consistent downstream evaluation, frozen pretrained models may differ in their sensitivity to input statistics, and future work should investigate whether the observed trends remain consistent when each pretrained model is evaluated using its native preprocessing pipeline.

Finally, interpretation of the domain-specific MAE results should be made with caution. The comparison with generalist foundation models is not strictly controlled, as the domain-specific model differs not only in pretraining data but also in architecture and pretraining scale. Consequently, the observed performance differences cannot be attributed solely to domain-specific pretraining. Nevertheless, the domain-specific MAE model achieved reasonable performance on the binary tumour classification task and demonstrated more structured attention patterns than the ImageNet-pretrained MAE encoder, suggesting that retinal-domain pretraining may influence the spatial organisation of patch representations. Future work should investigate larger-scale domain-specific foundation models with comparable architectures to enable a fairer evaluation of domain-specific versus generalist pretraining.

\section{Conclusion}
In this study, we investigated how different foundation model representations transfer to high-resolution UWF retinal image analysis within a controlled patch-based MIL framework. By preserving local retinal information through patch-based feature extraction and attention-based aggregation, the proposed framework enabled a systematic comparison of supervised and self-supervised Vision Transformer representations while avoiding the computational challenges associated with processing full-resolution UWF images.

Our results suggest that preserving high-resolution spatial information is beneficial for UWF retinal image analysis, with a supporting representation-strategy analysis favouring patch-based attention pooling over a single low-resolution global representation. Furthermore, the choice of pretraining objective substantially influenced downstream performance. Across the evaluated tasks, discriminative pretraining approaches, including supervised learning and self-distillation, produced more transferable feature representations than reconstruction-based Masked Autoencoder pretraining when used as frozen feature extractors. Attention analysis further showed that these representations resulted in distinct aggregation behaviours within the MIL framework, while partial fine-tuning substantially improved the performance of MAE-based models, highlighting the importance of downstream adaptation for reconstruction-based representations.

Overall, these findings demonstrate that the effectiveness of high-resolution UWF retinal image analysis depends not only on preserving spatial information, but also on the transferability of the underlying pretrained representations. The proposed framework provides a controlled approach for evaluating foundation model representations and offers practical insight into the selection of pretrained encoders for high-resolution retinal image analysis.

\appendices
\section{Domain-specific MAE Training}

\setcounter{table}{0}
\begin{table*}[!t]
\caption{Comparison of performance metrics for five-class DR classification on MMRDR across feature-pooling strategies for ViT-MAE and ViT-DINOv3 encoders. Results are based on a single training run. The best result within each encoder group is shown in bold.}
\label{tab:supplementary-pooling}
\centering
\scriptsize
\setlength{\tabcolsep}{4pt}
\begin{tabular}{@{}lcccccc@{}}
\toprule
\textbf{Model and pooling strategy} & \textbf{QWK} & \textbf{AUROC} & \textbf{F1} & \textbf{ACC} & \textbf{Precision} & \textbf{Recall} \\
\midrule
\multicolumn{7}{l}{\textit{ViT-MAE (ImageNet)+MIL}} \\
\quad CLS token & 0.4258 & 0.6922 & 0.3011 & 0.3589 & 0.3000 & 0.3589 \\
\quad Average pooling & \textbf{0.5233} & \textbf{0.7170} & \textbf{0.3412} & \textbf{0.4109} & \textbf{0.3088} & \textbf{0.4109} \\
\midrule
\multicolumn{7}{l}{\textit{ViT-DINOv3 (LVD)+MIL}} \\
\quad CLS token & 0.8633 & 0.8984 & 0.6445 & 0.6404 & \textbf{0.6680} & 0.6404 \\
\quad Average pooling & \textbf{0.8729} & \textbf{0.9016} & \textbf{0.6471} & \textbf{0.6681} & 0.6523 & \textbf{0.6681} \\
\bottomrule
\end{tabular}
\end{table*}

To investigate whether retinal-domain pretraining provides additional benefit over generalist foundation models, we trained a domain-specific Vision Transformer using the Masked Autoencoder (MAE) framework on unlabelled UWF Optomap images. This section describes the pretraining procedure. 

The foundation model was pretrained on a subset of the AlzEye 2018 dataset [33]. This dataset comprises 6,261,931 retinal images across seven modalities from 154,830 patients, with images from patients with glaucoma, pDR, cataract, or neovascular age-related macular degeneration. Within this dataset, there is a smaller subset of red/green UWF scanning laser ophthalmoscopy (SLO) images that we used for this study. The full dataset size was approximately 100,000 UWF Optomap images and training/validation/test set were split into 8:1:1.

During preprocessing, images were converted to single-channel greyscale representations before being used for foundation model training and for downstream tasks. Preliminary experiments comparing single-channel and two-channel inputs showed that greyscale representations produced lower validation loss during MAE pretraining. Using single-channel inputs also reduces memory usage, which is beneficial when training on large datasets. The resulting single-channel images were normalised with mean and standard deviation of 0.5 per channel.

Full images were downsampled to 1024×1024, then 8 patches of 224×224 pixels were randomly sampled from each image as model input.  By choosing a smaller number of sampled patches, we increased the variability across training samples. Patches with a black content of over 30\% were discarded.

We trained a Vision Transformer Small (ViT-S) [4] foundation model using the Masked Autoencoder (MAE) framework for self-supervised representation learning [5]. ViT-S was selected to balance representation capacity and computational feasibility given the dataset size and available hardware. This choice is also consistent with prior ophthalmic foundation models such as RETFound-Green [34], which used a ViT-S architecture when trained on approximately 70,000 retinal images.

Our ViT-S consisted of 12 Transformer blocks with an embedding dimension of 384 and token sizes of 16×16×1. The models divided the input images into non-overlapping image tiles and encoded them into 384-dimensional vectors and combined with fixed 2D sine-cosine positional encodings. These embeddings are processed through the Transformer blocks comprising multi-head self-attention [35] and feed-forward layers. 

The foundation model was trained from random initialisation using an AdamW optimiser with an initial learning rate of $1 \times 10^{-5}$ and a learning rate warmup of 5 epochs. The training objective was the mean squared error reconstruction loss with a mask ratio of 0.75. Mixed precision training was enabled to improve computational efficiency. Batch size was set to 128 and number of workers 8. The model was trained with a NVIDIA GeForce GTX TITAN X and trained for a maximum of 50 epochs, with early stopping (patience of 5 epochs based on validation loss). Model checkpoints achieving lowest validation losses were saved.

\begin{figure}[!t]
\centering
\includegraphics[width=\columnwidth]{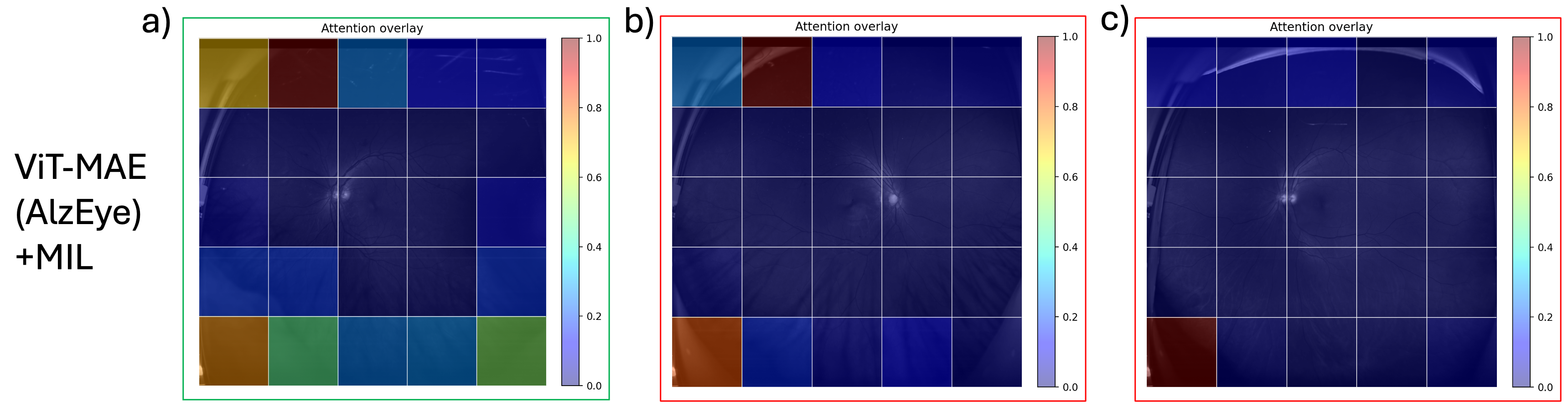}

\caption{Example attention heatmaps for ViT-MAE(AlzEye)+MIL model using three images from the evaluation set: (a) referable DR, (b) non-referable DR, and (c) non-referable DR. Green and red borders indicate correct and incorrect classifications, respectively.}
\label{fig:mae-attention-heatmaps}
\end{figure}

\section{Implementation Details}
All Vision Transformer encoders were instantiated from the \texttt{timm} library (v1.0.27) \cite{b36}. The MAE, DINOv1, and DINOv3 encoders used the \texttt{timm} checkpoints \path{vit_base_patch16_224.mae}, \path{vit_base_patch16_224.dino}, and \path{vit_base_patch16_dinov3}, respectively. The supervised ViT-B and ResNet baselines were loaded from \texttt{torchvision} \cite{b37}. The \path{vit_b_16}, \texttt{resnet50}, and \texttt{resnet18} models used \path{IMAGENET1K_V1} weights. Patch features were extracted using CLS-token pooling for the supervised and DINOv1 encoders and average pooling over patch tokens for the MAE and DINOv3 encoders, as detailed in Appendix C. 
 
\section{Feature Pooling Strategies for Pretrained Vision Transformers}
Unlike standard ViT classification pipelines, the proposed MIL framework operates on patch-level representations rather than image-level predictions. Consequently, the choice of feature pooling strategy depends on the pretraining objective used by each encoder.

Because the MAE classification token is not directly optimised by the reconstruction objective, average pooling over spatial tokens was used in the main experiments, while CLS-token pooling was evaluated as a sensitivity analysis. ViT-Supervised and DINOv1 were evaluated using CLS-token representations, consistent with their pretraining objectives. For DINOv3, the main experiments retained the default average-pooling configuration of the timm implementation, and CLS-token pooling was subsequently evaluated in a post-hoc sensitivity analysis. The corresponding pooling comparisons are reported in Appendix Table I.

\section{Full Performance Metrics}
Appendix Tables II-IV provide the complete evaluation results corresponding to the experiments presented in the main manuscript, including all reported performance metrics.

\begin{table*}[!t]
\caption{Full performance metrics for five-class DR classification on MMRDR. Values are reported as mean $\pm$ standard deviation across three runs. Best results are shown in bold.}
\label{tab:full-performance-metrics}
\centering
\scriptsize
\setlength{\tabcolsep}{4pt}
\begin{tabular}{lcccc}
\hline
\textbf{Model} & \textbf{QWK} & \textbf{AUROC} & \textbf{F1} & \textbf{ACC} \\
\hline
ResNet-50(ImageNet)+MIL & $0.6907 \pm 0.0016$ & $0.7804 \pm 0.0049$ & $0.4098 \pm 0.0564$ & $0.4438 \pm 0.0622$ \\
ViT-MAE(AlzEye)+MIL & $0.3247 \pm 0.0474$ & $0.6301 \pm 0.0034$ & $0.2276 \pm 0.0509$ & $0.3201 \pm 0.0391$ \\
ViT-MAE(ImageNet)+MIL & $0.5325 \pm 0.0205$ & $0.7282 \pm 0.0097$ & $0.3563 \pm 0.0155$ & $0.4170 \pm 0.0156$ \\
ViT-Supervised(ImageNet)+MIL & $0.8396 \pm 0.0043$ & $0.8781 \pm 0.0026$ & $0.6019 \pm 0.0295$ & $0.6179 \pm 0.0006$ \\
ViT-DINOv1(ImageNet)+MIL & $0.8602 \pm 0.0034$ & $\mathbf{0.8963 \pm 0.0017}$ & $0.6204 \pm 0.0168$ & $\mathbf{0.6546 \pm 0.0070}$ \\
ViT-DINOv3(LVD)+MIL & $\mathbf{0.8629 \pm 0.0102}$ & $0.8958 \pm 0.0051$ & $\mathbf{0.6332 \pm 0.0244}$ & $0.6491 \pm 0.0166$ \\
\hline
\end{tabular}
\end{table*}

\begin{table*}[!t]
\caption{Full performance metrics for binary referable DR classification on DeepDRiD. Values are reported as mean $\pm$ standard deviation across three runs. Best results are shown in bold.}
\label{tab:deepdrid-full-performance}
\centering
\scriptsize
\setlength{\tabcolsep}{4pt}
\begin{tabular}{lcccc}
\hline
\textbf{Model} & \textbf{AUROC} & \textbf{AUPR} & \textbf{ACC} & \textbf{Kappa} \\
\hline
ResNet-18(ImageNet)+MIL & $0.6520 \pm 0.1576$ & $0.6723 \pm 0.1981$ & $0.6026 \pm 0.1094$ & $0.2087 \pm 0.1953$ \\
ViT-MAE(AlzEye)+MIL & $0.5207 \pm 0.0353$ & $0.4738 \pm 0.0465$ & $0.5000 \pm 0.0000$ & $0.0451 \pm 0.0108$ \\
ViT-MAE(ImageNet)+MIL & $0.8131 \pm 0.0053$ & $0.7920 \pm 0.0061$ & $0.7692 \pm 0.0000$ & $0.5348 \pm 0.0032$ \\
ViT-Supervised(ImageNet)+MIL & $0.8520 \pm 0.0442$ & $0.8128 \pm 0.0623$ & $0.7628 \pm 0.0400$ & $0.5169 \pm 0.0827$ \\
ViT-DINOv1(ImageNet)+MIL & $0.8136 \pm 0.0616$ & $0.8013 \pm 0.0625$ & $0.7692 \pm 0.0333$ & $0.5178 \pm 0.0660$ \\
ViT-DINOv3(LVD)+MIL & $\mathbf{0.9030 \pm 0.0244}$ & $\mathbf{0.8823 \pm 0.0250}$ & $\mathbf{0.8526 \pm 0.0222}$ & $\mathbf{0.6969 \pm 0.0503}$ \\
\hline
\end{tabular}
\end{table*}

\begin{table*}[!t]
\caption{Full performance metrics for binary intraocular tumour classification. Values are reported as mean $\pm$ standard deviation across five folds. Best results are shown in bold.}
\label{tab:tumour-full-performance}
\centering
\scriptsize
\setlength{\tabcolsep}{4pt}
\begin{tabular}{lcccc}
\hline
\textbf{Model} & \textbf{AUROC} & \textbf{AUPR} & \textbf{ACC} & \textbf{Kappa} \\
\hline
ResNet-18(ImageNet)+MIL & $0.8478 \pm 0.1370$ & $0.8533 \pm 0.1463$ & $0.7909 \pm 0.0951$ & $0.5819 \pm 0.1902$ \\
ViT-MAE(AlzEye)+MIL & $0.8613 \pm 0.0101$ & $0.8826 \pm 0.0071$ & $0.7916 \pm 0.0170$ & $0.5833 \pm 0.0341$ \\
ViT-MAE(ImageNet)+MIL & $0.8742 \pm 0.0199$ & $0.9005 \pm 0.0192$ & $0.8204 \pm 0.0315$ & $0.6409 \pm 0.0629$ \\
ViT-Supervised(ImageNet)+MIL & $0.9806 \pm 0.0092$ & $0.9824 \pm 0.0070$ & $0.9359 \pm 0.0162$ & $0.8719 \pm 0.0324$ \\
ViT-DINOv1(ImageNet)+MIL & $\mathbf{0.9963 \pm 0.0011}$ & $\mathbf{0.9965 \pm 0.0012}$ & $\mathbf{0.9742 \pm 0.0026}$ & $\mathbf{0.9485 \pm 0.0052}$ \\
ViT-DINOv3(LVD)+MIL & $0.9894 \pm 0.0037$ & $0.9902 \pm 0.0031$ & $0.9492 \pm 0.0137$ & $0.8984 \pm 0.0273$ \\
\hline
\end{tabular}
\end{table*}



\EOD

\end{document}